\documentclass{article}

\PassOptionsToPackage{numbers}{natbib}

\usepackage[preprint]{neurips_2021}

\usepackage[utf8]{inputenc} 
\usepackage[T1]{fontenc}    
\usepackage{hyperref}       
\usepackage{url}            
\usepackage{booktabs}       
\usepackage{amsfonts}       
\usepackage{nicefrac}       
\usepackage{microtype}      
\usepackage{xcolor}         
\usepackage{multirow}
\usepackage{graphicx}
\usepackage{float}
\usepackage{subcaption}
\usepackage{diagbox}
\usepackage{todonotes}

\hypersetup{colorlinks=true}
\hypersetup{linkcolor=blue!10!black}                                            
\hypersetup{urlcolor=red!20!black}
\hypersetup{citecolor=green!40!black}

\title{HTTP2vec: Embedding of HTTP Requests for Detection of Anomalous Traffic}

\author{%
  Mateusz~Gniewkowski\\
  Faculty of Electronics\\
  Wroclaw University of Science and Technology\\
  Poland \\
  \texttt{mateusz.gniewkowski@pwr.edu.pl} \\
  \And
  Henryk~Maciejewski \\
  Faculty of Electronics\\
  Wroclaw University of Science and Technology\\
  Poland \\
  \texttt{henryk.maciejewski@pwr.edu.pl} \\
  \And
  Tomasz~Surmacz \\
  Faculty of Electronics\\
  Wroclaw University of Science and Technology\\
  Poland \\
  \texttt{tomasz.surmacz@pwr.edu.pl} \\
  \And
  Wiktor~Walentynowicz \\
  Faculty of Computer Science and Management\\
  Wroclaw University of Science and Technology\\
  Poland \\
  \texttt{wiktor.walentynowicz@pwr.edu.pl} \\
}

\begin{document}

\maketitle

\begin{abstract}
  Hypertext transfer protocol (HTTP) is one of the most widely used protocols on
the Internet. As a consequence, most attacks (i.e., SQL injection, XSS) use
HTTP as the transport mechanism. Therefore, it is crucial to develop an
intelligent solution that would allow to effectively detect and filter out
anomalies in HTTP traffic. Currently, most of the anomaly detection systems
are either rule-based or trained using manually selected features. We propose
utilizing modern unsupervised language representation model for embedding HTTP
requests and then using it to classify anomalies in the traffic.  The solution
is motivated by methods used in Natural Language Processing (NLP) such as
Doc2Vec which could potentially capture the true understanding of HTTP messages,
and therefore improve the efficiency of Intrusion Detection System.  In our
work, we not only aim at generating a suitable embedding space, but also at the
interpretability of the proposed model. We decided to use the current
state-of-the-art RoBERTa, which, as far as we know, has never been used in a
similar problem.  To verify how the solution would work in real word
conditions, we train the model using only legitimate traffic. We also try to
explain the results based on clusters that occur in the vectorized requests
space and a simple logistic regression classifier. We compared our approach
with the similar, previously proposed methods. We evaluate the feasibility of our
method on three different datasets: CSIC2010, CSE-CIC-IDS2018 and one that we
prepared ourselves. The results we show are comparable to others or better, and
most importantly -- interpretable.




\end{abstract}

\section{Introduction}
\label{section:intro}

According to OWASP Top Ten, injection attacks are number one threat in the modern Internet, but many others are also very common, such as Cross-Site Scripting, password or content brute-forcing, or exploiting server misconfigurations. Attack vectors targeting Web servers naturally utilize HTTP protocol as a transport mechanism, so it is essential to develop solutions that would allow not only to detect anomalies in the protocol requests, but also to help with potential post incident analysis.

When it comes to detecting attacks, there are two possible strategies: pattern detection
or trained models (supervised or unsupervised anomaly detection).
Within the context of text protocols like HTTP, pattern matching is based on expert knowledge and sets of rules (patterns) that if matched, indicate an attack.
For example, if the document contains multiply percent-encoded characters, it could mean that someone is trying to bypass an input sanitation. On that principle Snort tool is built (at least partially).

If one would like to apply trained models to solve the problem, the most important task is a feature construction.
Over the years, many text vectorization or feature learning methods have been developed,
including simple ones, like \emph{bag-of-words}, \emph{tf-idf} or \emph{bag-of-n-grams}; and more advanced like \emph{Doc2Vec}, \emph{fasText}, \emph{ELMo} or \emph{BERT}. Some of them have already been successfully used for solving the anomaly detection problem in HTTP traffic.




\citet{torrano2015combining} proposed a solution which combines expert knowledge with n-gram feature construction. They also use several decision tree algorithms as classifiers. In \citep{li2020weighted} authors propose a combination of tfidf and word2vec for generating vectors and then apply gradient boosting to classify them. \citet{vartouni2018anomaly}  proposed another n-gram based model that also utilizes autoencoder neural network to further reduce the dimensionality of the data. On the similar principle \citet{das2020network} developed a solution based on Doc2Vec method.

Some of the works, rather than generating the vectors first and then using them in a classification,
present deep learning approach which on the output determine the type of label. In other words they are fully supervised and need labels to establish weights in hidden layers. Such methods include, among others, \citep{liu2019deep} (deep neural network based on LSTM and CNN layers) and \citep{luo2020novel} (combines intermediate vectors obtained from CNN, LSTM and MRN model).

In our work we focus on a method that allows obtaining a vector space for classification in unsupervised manner.
We believe that this approach is more suitable in real-word scenarios as it does not require any labelling (the vectors can be still classified by any outlier detection algorithm).
Moreover, we decided to train the model using only normal traffic, to check if the used classifier would still be able to detect anomalies.
Our main goal was to build an HTTP requests space that behaves well (points are well separated and require little effort for classification) and to show how the obtained space can be used by an expert as a tool for analyzing traffic.

The main contributions are the following. We propose a method for detecting and analyzing anomalous HTTP traffic. The main idea is to use the RoBERTa model \citep{liu2019roberta} to obtain vector representations of HTTP requests. We show that using this representation, we can achieve state-of-the-art performance in the task of supervised detection of anomalous HTTP requests.  We also show how the vector representations can be used to analyze HTTP traffic and identify interpretable patterns of tokens which are specific to the anomalies discovered in HTTP traffic. It is worth noting that the interpretability of results is an important and unique characteristic of the proposed method. Finally, we show that using the proposed vector representations we can cluster and visualize patterns in HTTP traffic. 
We performed a feasibility study of the proposed method using CSIC2010, CSE-CIC-IDS2018 benchmark datasets, and another one that we prepared ourselves.

\section{Related works}
\label{section:related}


The HTTP request analysis can be represented as a Natural Language Processing problem (or rather Not Natural Language Processing problem, as it is an artificial language). The most important task we face is choosing a language model that would allow to obtain a vector space. In our work we decided to utilize RoBERTa \citep{liu2019roberta} model (described in detail in the next section) as it is the current state-of-the-art solution for plenty of down-stream tasks in NLP. The model not only resolves the Out of Vocabulary problem (OOV), but also, thanks to self-attention mechanism, better encodes tokens based on their context (e.g. imagine how many things can be represented by a ``/'' in HTTP request).

The most related works to our paper are those that also use CSIC2010 dataset (this is probably the only dataset that contains HTTP requests in a ready to use text form) and focus on unsupervised learning representation. Sadly, most of them define their experiments differently, as they use anomalous traffic in the process of generating vectors.


Work based on a similar idea to ours is a solution based on the Doc2Vec vectorization method~\citep{das2020network}. HTTP traffic is processed by the Doc2Vec model into a vector form, which is then used to predict whether the traffic is anomalous or normal. The Doc2Vec model is trained on the entire CSIC2010 set (both normal and anomalous traffic -- the combined training and test sets). HTTP requests are modified to represent the first line of that request and grouped by 10 into "documents". This way, each document represents either 10 normal or 10 anomalous requests. Classification is performed by an ensemble of classifiers trained on the training data, which represents 70\% of the total data, and tested on the test data, which represents the remaining 30\%.

Next work, by \citet{vartouni2018anomaly}, represents HTTP queries as bigrams on a dictionary of 80 ASCII characters. This yielded 2572 features representing the HTTP query. An autoencoder model was used to learn the representation for the classifier. The Isolation Forest algorithm was used to determine the association of a given HTTP query on the obtained vector space to either normal or anomalous traffic. Despite performing experiments on the familiar CSIC2010 dataset, there is a noticeable difference in the reported data counts for the datasets, so we assume that a smaller subset of the data was used.

The next work we found presents a fully supervised approach \citep{liu2019deep}. An LSTM-CNN neural architecture is used to classify HTTP traffic. First, the LSTM recurrent network processes the HTTP query based on block-based features, then the selected states of the LSTM network are fed to the convolutional network, which, after processing the vectors, passes them to the output in the form of MLP network which classifies the HTTP query into one of two classes. The authors of the method report very good results not only for the CSIC2010 collection, but also for CICIDS 2017 and ISCX 2012 collections containing different types of attacks.

In \citep{luo2020novel} there is a description of a web attack detection system that is based on an ensemble of classifiers and vector representation techniques from the NLP domain. This system first tokenizes text based on a manually prepared dictionary containing tokens characteristic for network traffic. The resulting text representations are vectorized in parallel using neural models based on recurrent and convolutional networks. A comprehensive check is then performed on them, which returns an evaluation vector specifying the output confidence of the vectors on their mutual difference. The produced vector and the vectors from the neural models go to an ensemble of classifiers, which evaluates whether the HTTP query is a normal traffic or an attack. The method was tested on the CSIC2010 collection and own collections, unfortunately there is no information on how the model training was performed.





We chose to reproduce the work of \citep{park2018anomaly} because of the similarity of the idea. The authors of the paper designed a convolutional autoencoder that learns to reconstruct an HTTP message transformed to a character-binary image. The CAE (convolutional autoencoder) consists of consecutive convolutional layers, of which the main part is the layers based on the Inception-ResNet-v2 architecture. CAE includes a decoding part that is designed as an inverted encoder. The input data of the network are character-binary images obtained by transforming the text into a matrix of occurrences of characters from the dictionary (68 values) at a given position throughout the HTTP message. The network learns to reproduce such a representation only on normal HTTP messages with no anomalies, and the learning criterion 
BCE (binary cross entropy). Anomaly detection is supposed to be based on showing a difference in the reconstructed representation under the assumption that HTTP messages containing anomalies will have a varying BCV (\emph{binary cross varentropy}) value.

When reproducing this solution, we encountered problems, such as mismatches between the design possibilities and the diagrams in the paper.
In case of the stem layers in the encoder and decoder, we had to adjust the convolution layer due to a mismatch in size, and correct the convolution layer in the Reduction-B module. Also, in some places we had to adjust the padding window size ourselves, due to the lack of description of these values.

\section{Proposed approach}
\label{section:proposed}
\begin{figure}
    \centering
    \includegraphics[width=0.95\textwidth]{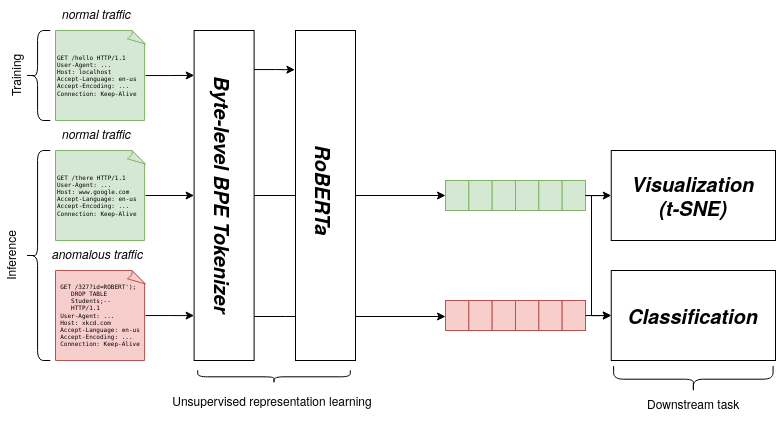}
    \caption{The HTTP2vec pipeline}
    \label{fig:http2vec}
\end{figure}

\subsection{Tokenization}

Byte Pair Encoding (BPE) is a compression method of replacing the most common byte pairs in a data by a byte that does not occur in that data.
BPE has been transferred to the field of natural language processing as a method of text tokenization \citep{sennrich2015neural}.  It focuses on grouping the most frequent character strings in the learning corpus.
The method starts at the level where a single character is a single token, first grouping pairs of characters, then triplets and so on, until a dictionary is created which contains the number of tokens preset for the model.
An extension method of BPE is Byte-level Byte Pair Encoding (BBPE) \citep{wang2020neural}, which is based on a dictionary of bytes rather than characters. This allows the dictionary to be kept small while recognizing many different forms.

For the purpose of the tasks described in our paper, the tokenizer is based on BBPE and is trained on the entire dataset -- containing both normal and anomalous traffic. This allows us to create a reasonably accurate tokenizer and avoid the pitfall of biasing the anomaly detection method by detecting a large number of short tokens after tokenization, which would not necessarily mean anomalous traffic in reality. The implementation we use is based on the solution provided by HuggingFace Transformers~\citep{wolf-etal-2020-transformers}. The output of our tokenizer is the input of the RoBERTa model.



\subsection{Language model}

Our method is based on vectorization using the RoBERTa model \citep{liu2019roberta}, which in turn is based on the BERT model \citep{devlin2018bert}.
Bidirectional Encoder Representations from Transformers (BERT) is a language model built on the Transformer architecture.
It uses an attention mechanism and allows to process data sequentially, considering information about the token position in the sequence without using recursion.
The key element of BERT is the inclusion of two-sided context for each token of the processed sequence.
The standard Transformer architecture distinguishes between an encoder and decoder part. However, in case of BERT, we are mainly interested in the  encoder's output.
Unlike standard sequence processing methods, the Transformer architecture processes all tokens simultaneously, not just in a selected direction.
The input of the method is a sequence of tokens, which are first embedded in a vector space and then processed by the neural model. The learning process of the BERT model is based on two methods: Masked Language Model (MLM) and Next Sentence Prediction (NSP).

In MLM method, a given percentage of tokens is hidden via a [MASK] token, and then the network attempts to predict the contents of the hidden token (this requires the connection of the classifier).
The NSP method is based on the network receiving two sentences as input, and the network's task is to determine if the second sentence is a sentence following the first sentence.
These two learning methods can be combined.

A modification of the BERT model is RoBERTa (\emph{Robustly optimized BERT approach}). It improves the original method in relation to model learning. The authors of RoBERTa emphasize four improvements they have implemented:
\begin{itemize}
    \item Dynamic Masking
    \item Full-sentences without NSP loss
    \item Large mini-batches
    \item Larger BBPE
\end{itemize}
The main change is to process as large parts of the document as possible without paying attention to sentence boundaries, instead of processing single sentences and to exclude the NSP task from learning.
BBPE was also used instead of BPE as a tokenization method.
Dynamic masking minimally improved the performance of the model.

The RoBERTa model that we use in our method is based on the implementation available in the HuggingFace Transformers \citep{wolf-etal-2020-transformers} package. We train the model only on training data representing HTTP messages without anomalies.
The chosen size of the model is ``base'' with a maximum sequence length of 512, number of layers set to 6, and 12 attention heads.
We trained the model for 10 epochs with a batch size of 32, with 15\% masked tokens. We use this learned model to generate vectors of length 3072 ($4\cdot768$) -- the concatenation of the last 4 layers. The final vector for the HTTP message is the average of all message lines with the same vector length.





\subsection{Classification}

In order to classify the obtained vectors, one can use any suitable algorithms. In this work we decided to utilize the following three algorithms, as we believe that they provide a good intuition on how the embedding space looks like: 1) Logistic Regression (LR), 2) Random Forest (RF), 3) Support Vector Classification (SVC) with linear kernel. 
In our experiments we simply used Scikit-learn \citep{scikit-learn} implementations with most of the parameters left as defaults (although we increased a max number of iterations in RL algorithm to 500).

\subsection{Visualization}

For visualization of the obtained embeddings on a 2D plane we decided to use t-SNE \cite{van2008visualizing} for dimensionality reduction. The method is believed
to preserve global structure better than classical multidimensional scaling \cite{MDS}, because it preserves the similarity between points defined as normalised Gaussian. Therefore, it uses Euclidean distance in the original space.
The similarities in low dimensional space are model by normalised Student-t distribution. 
The t-SNE method minimises the Kullback–Leibler divergence between the similarities in both spaces with respect to the locations of the points in the low-dimensional space.

\section{Experiments}
\label{section:experiments}
\subsection{Datasets}

Every dataset we used was divided into two parts: the training part (normal requests, used only for representation learning) and the inference part (normal and anomalous traffic encoded via the model and used for classification). The sizes of individual sets are shown in table \ref{tab:datasets}. We briefly describe the datasets below.

\paragraph{CSIC2010}

The HTTP dataset CSIC2010%
\footnote{CSIC2010 is available at \url{https://www.tic.itefi.csic.es/dataset/}}%
\cite{gimenez2010http}
is the dataset most often used in similar problems, as it contains ready-to-use text files with HTTP requests. 
The dataset includes attacks such as SQL injection, buffer overflow, information gathering, files disclosure, CRLF injection, XSS, and few other attacks. Authors generated the traffic by targeting a single e-Commerce web application, which makes it perfect for utilizing in our approach. Due to the need to detect CRLF injection attacks, we decided to encode CR (carriage return) an LF (line feed) characters as literal ``\textbackslash r'' and ``\textbackslash n'' strings.

\paragraph{CSE-CIC-IDS2018}

CSE-CIC-IDS2018\footnote{CSE-CIC-IDS2018 is available at \url{https://registry.opendata.aws/cse-cic-ids2018/}} \cite{sharafaldin2018toward} is a well-known dataset developed by Communications Security Establishment and the Canadian Institute for Cybersecurity. Despite its excellent quality, it was not designed for the problem discussed in this paper. Luckily, it contains several web-oriented attacks like ``Brute Force -- Web'', ``Brute Force -- XSS'' and ``SQL Injection'' that we extracted from captured packets (we used Friday traffic, 23-02-2018). This dataset differs from the others in that the normal traffic requests are directed to many web applications. In order to generate anomalous traffic, the authors used DVWA application\footnote{https://dvwa.co.uk/} that was hosted on a single machine. This forced us to make some minor changes in the requests in order to avoid spurious correlations:

\begin{enumerate}
    \item Every ``Host'' field in the requests always pointed to a single IP address. We decided to randomly change that to any value that appeared in the normal dataset.
    \item Every URI in the request always started with one of the following strings: ``/DVWA/vulnerabilities/xss'', ``DVWA/dvwa'', ``/DVWA''. We simply removed those.
    \item Every anomalous request contained ``Upgrade-Insecure-Requests'' header, which we also removed.
\end{enumerate}

Also, the normal traffic not always contained text payloads, so we removed those that did not match a proper ``Content-Type'' (i.e. ``application/json'').

\paragraph{UMP}

UMP dataset\footnote{The web application used for collecting the data: \url{http://trasy.ump.waw.pl/}} is one, that we prepared by ourselves.
It is based on a web application (provided in the footnote) that provides a detailed map of Poland and surrounding countries and allows route planning. Thanks to the fact, that one of us (\emph{anonymized name}) is responsible for hosting the web page, we were able not only to collect a legitimate traffic from real users, but also to generate the anomalous requests as well. For this purpose  we decided to use an excellent tool called Arachni\footnote{\url{https://www.arachni-scanner.com/}} (using the default options to execute the scan). The generated attack data had very similar headers and nearly no payloads, so we decided to use only the first lines of HTTP requests to avoid oversimplification. 
    
{
\small
\begin{table}[htb!]
\centering
\caption{The division of the datesets used in experiments. The training part is only used for RoBERTa model. The inferece parts are used in classification stage.}
\label{tab:datasets}
\begin{tabular}{cccc}
\hline
\multirow{2}{*}{Division} & \multicolumn{3}{c}{Datasets}       \\ \cline{2-4} 
                              & CSIC2010 & CSE-CIC-IDS2018 & UMP    \\ \hline
normal (train)                & 36000    & 591175          & 150156 \\ \hline
normal (inference)                 & 36000    & 13591           & 35876  \\ \hline
anomaly (inference)                & 25065    & 13591           & 35876  \\ \hline
\end{tabular}
\end{table}
}

\subsection{Baselines}

In order to compare with other approaches we have selected several works that
also used the CSIC2010 dataset. We have described all of them in more detail in Section
\ref{section:related}

\citet{vartouni2018anomaly} conducted experiments similar to ours, although
it seems that they used only a subset of the CSIC2010 dataset. Their solution
is fully unsupervised as they decided to utilize Isolation Forest algorithm for
classification. The article reports 84.12\% of F1-score, which is the minimum of what we want
to achieve (unsupervised classification methods tend to perform worse than
supervised).

\citet{liu2019deep} achieved 99.12\% of TPR and 0.22\% of FPR score (FPR99 = 0.22\%). Their
solution is fully supervised, so we will probably get slightly worse results.
However, it is worth noting that their test set is smaller and more unbalanced
than ours -- this could lead to a slight overestimation of the results
(especially TPR). We treat their results exemplary.

At this point, it is also appropriate to mention a paper \citep{das2020network}
as their authors present the most similar approach to ours. Its authors
present results close to a perfect classifier, but also made some assumptions
we consider wrong. First, the language model (Doc2Vec) is trained on a whole
dataset (same samples are used for training and inference). Secondly, they
combined ten different samples into one document, which makes the problem
much easier to solve. Generally using Doc2Vec as a language model is not a bad
idea, but RoBERTa is proven to perform better in multiple down-stream tasks in
NLP. Moreover, Doc2Vec does not solve the Out of Vocabulary problem.

In our experiments we also tried to recreate the CAE architecture from \cite{park2018anomaly} using the hyperparameters given by the authors. The method converged to a local minimum in the second epoch and did not learn further. Based on our failed experimental attempts described in Section \ref{section:related}, we conclude that their results are irreproducible without access to the original source code.

%

\subsection{Results}

\begin{table}[htb!]
\footnotesize
\centering
\caption{Performance comparisons of the proposed vectorization method (based on RoBERTa) and CAE using several classification algorithms. All the results come from the stratified k-folds cross-validation strategy with k=5.}
\label{tab:results}
\begin{tabular}{cccccc}
\hline
Dataset                             & Method & FPR90                & FPR99                & F1                  & MCC \citep{matthews1975comparison}                \\ \hline
\multirow{3}{*}{CSIS2010}           & LR     & $0.017 \pm 0.004$   & $0.075 \pm 0.006$    & $0.951 \pm 0.004$    & $0.916 \pm 0.006$   \\ \cline{2-6}
                                    & SVC    & $0.003 \pm 0.001$   & $0.042 \pm 0.005$    & $0.969 \pm 0.002$    & $0.948 \pm 0.003$   \\ \cline{2-6}
                                    & RF     & $0.010 \pm 0.002$   & $0.070 \pm 0.006$    & $0.959 \pm 0.002$    & $0.930 \pm 0.004$   \\ \cline{1-6}
\multirow{3}{*}{UMP}                & LR     & $0.120 \pm 0.000$   & $0.150 \pm 0.000$    & $0.926 \pm 0.001$    & $0.850 \pm 0.002$   \\ \cline{2-6}
                                    & SVM    & $0.120 \pm 0.002$   & $0.153 \pm 0.003$    & $0.926 \pm 0.001$    & $0.850 \pm 0.002$   \\ \cline{2-6}
                                    & RF     & $0.120 \pm 0.000$   & $0.150 \pm 0.002$    & $0.926 \pm 0.001$    & $0.850 \pm 0.002$   \\ \cline{1-6}
\multirow{3}{*}{CSE-CIC-IDS2018}    & LR     & $0.000 \pm 0.000$   & $0.000 \pm 0.000$    & $0.999 \pm 0.001$    & $0.998 \pm 0.001$   \\ \cline{2-6}
                                    & SVM    & $0.000 \pm 0.000$   & $0.000 \pm 0.000$    & $0.999 \pm 0.001$    & $0.998 \pm 0.001$   \\ \cline{2-6}
                                    & RF     & $0.000 \pm 0.000$   & $0.000 \pm 0.000$    & $0.999 \pm 0.001$    & $0.998 \pm 0.001$   \\ \hline
\end{tabular}
\end{table}


\begin{figure}[htb!]
\centering
   \begin{subfigure}[b]{0.32\linewidth} \centering
     \includegraphics[width=\textwidth]{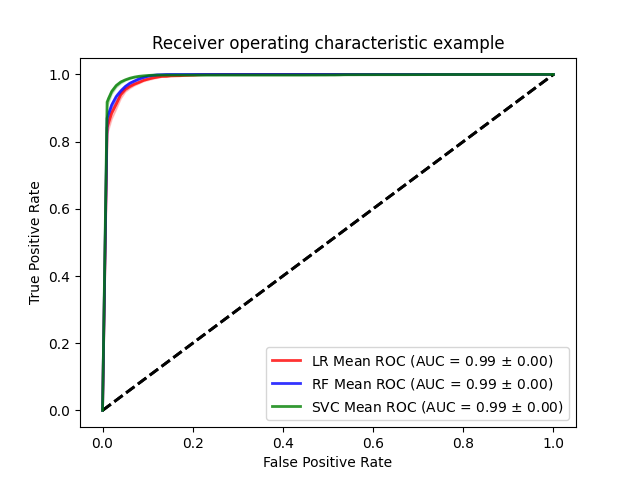}
     \caption{CSIC2010}\label{fig:figAA}
   \end{subfigure}
   \begin{subfigure}[b]{0.32\linewidth} \centering
     \includegraphics[width=\textwidth]{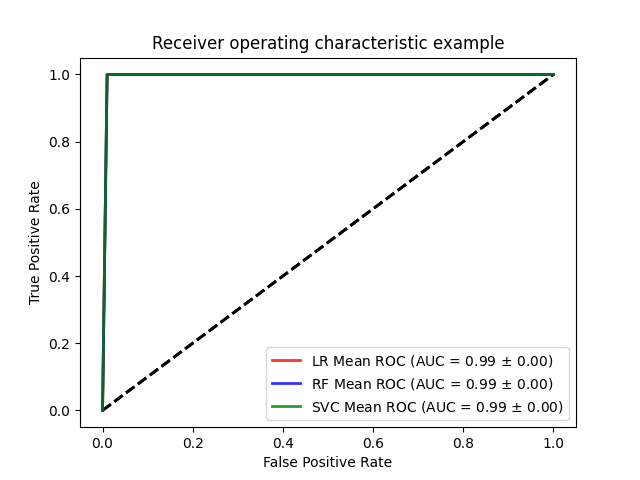}
     \caption{CSE-CIC-IDS2018}\label{fig:figBB}
   \end{subfigure}
   \begin{subfigure}[b]{0.32\linewidth} \centering
     \includegraphics[width=\textwidth]{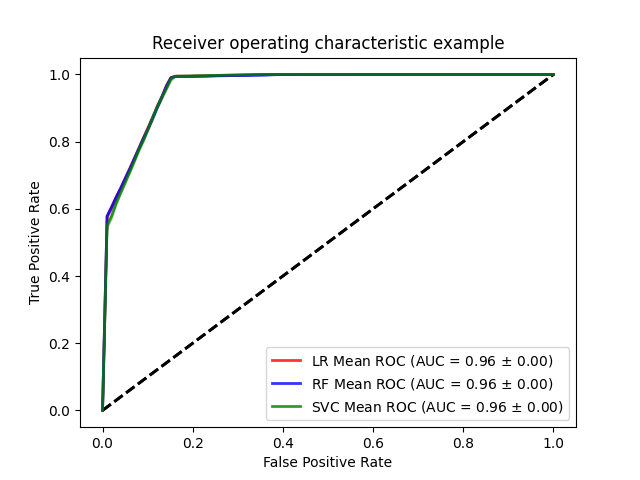}
     \caption{UMP}\label{fig:figCC}
   \end{subfigure}
\caption{Mean (K-Folds, $k=5$) ROC curves for the HTTP2vec approach.} \label{fig:rocs}
\end{figure}

After generating the vectors (the number of samples is shown in Table
\ref{tab:datasets}), we used them to train three different classifiers: LR, SVC
and RF (more details in Section \ref{section:proposed}). Note, that we split the data
using Stratified K-Folds algorithm (with k=5). The mean results together with
their standard deviation across all folds, are presented in Table
\ref{tab:results}.

The results for CSIC2010 dataset show that we easily managed to exceed our
baseline requirements. The best classifier turned out to be SVC with the F1-Score of
96.9\%. We decided to use a linear kernel, which means that the samples
are well-separated in higher dimension. In comparison with results we
considered ideal, we obtained FPR99 = 4.2\%, which is much worse than those
reported by~\cite{liu2019deep}. We believe, that this is mainly due to a
different usage of the data.

The results for CSE-CIC-IDS2018 dataset show that we obtained a perfect
classifier. This means that either the dataset is really simple to classify
or we made some mistakes during data processing.  We discussed this
problem in the next section.

The results for the UMP dataset (the one we created by ourselves) suggest that there is
some kind of subset of samples that is nearly impossible to classify. This
conclusion follows from the shape of ROC curve in Figure \ref{fig:rocs} and the
fact that all classifiers returned the same results. According to our knowledge
about the dataset, this is because the tool we used for creating anomaly subset,
actually generates a lot of legitimate traffic (scanning for HTTP headers used,
building the dependency tree for a webpage).

\section{Interpretability}
\label{section:interpretability}
\begin{figure}[htb!]
\centering
   \begin{subfigure}[b]{0.32\linewidth} \centering
     \includegraphics[width=\textwidth]{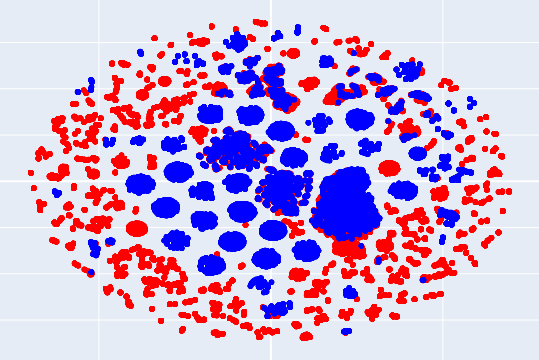}
     \caption{CSIC2010}\label{fig:figA}
   \end{subfigure}
   \begin{subfigure}[b]{0.32\linewidth} \centering
     \includegraphics[width=\textwidth]{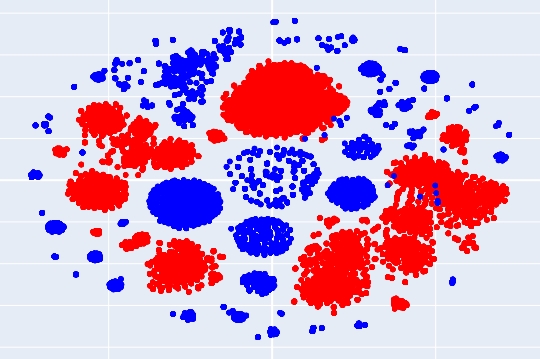}
     \caption{CSE-CIC-IDS2018}\label{fig:figB}
   \end{subfigure}
   \begin{subfigure}[b]{0.32\linewidth} \centering
     \includegraphics[width=\textwidth]{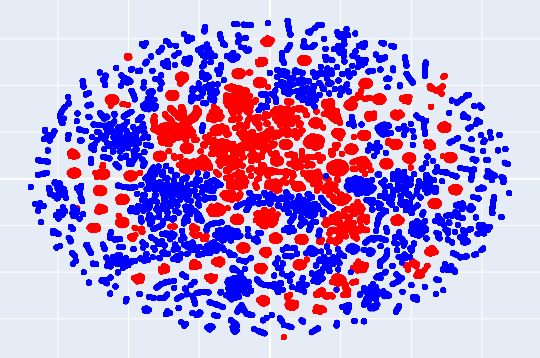}
     \caption{UMP}\label{fig:figC}
   \end{subfigure}
\caption{The vectorized requests space reduced to 2D with t-SNE. The red color represents anomalous traffic} \label{fig:tsne}
\end{figure}

Let us start with the result of t-SNE algoritm. The plots in Figure
\ref{fig:tsne} shows that classes are well separated even in a lower space.
This shows that similar HTTP requests are actually grouped together and brings 
the possibility of investigating a neighborhood of any sample. In Table
\ref{tab:simmilar} we show several n-th closest (using Euclidean distance) samples to a given one in a
first row. As one can see, SQL Injection attacks are actually close to each
other. Interestingly, further neighborhood of the sample contains other type of
injections -- operating system commands. We believe that this approach could be utilized by an expert to identify similar attack attempts and therefore help with post-incident analysis.

\begin{figure}[htb!]
    \centering
    \includegraphics[width=0.6\textwidth]{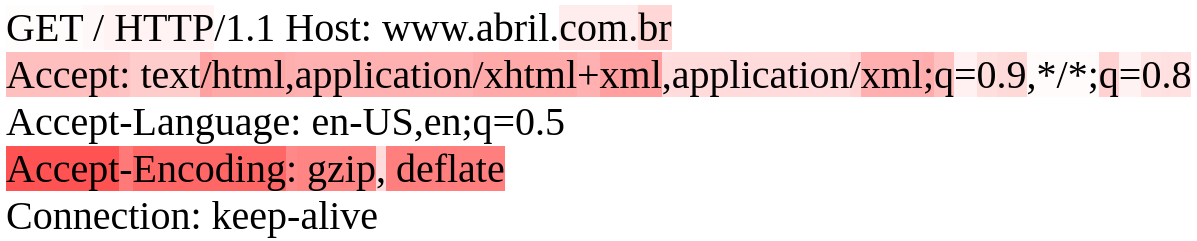}
    \caption{An example anomaly request from CSE-CIC-IDS2018 that shows most important features indicating anomality of the sample. The marked pattern repeats among all anomalous traffic, which makes the dataset easy to classify.}
    \label{fig:idsinterp-ids2018}
\end{figure}

\begin{figure}[htb!]
    \centering
    \includegraphics[width=0.9\textwidth]{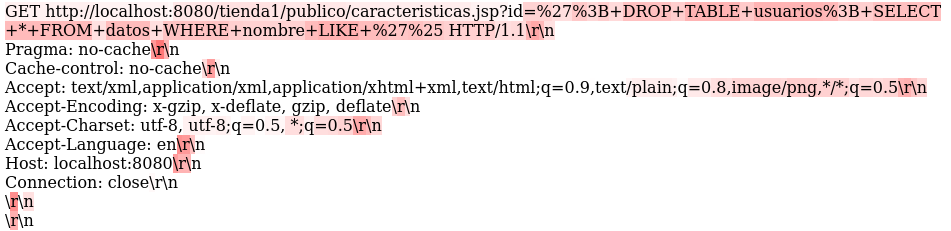}
    \caption{An example anomaly request from CSIC2010 that shows most important features indicating anomality of the sample. The marked pattern indicates that the anomaly is SQL Injection.}
    \label{fig:idsinterp-csic2010}
\end{figure}

This leads to a question: What features of the requests prove their anomaly? To find an answer, 
we first generated a list of tokens, then a set of documents which were based on the original one, but with one token removed from each of them. Then we generated vectors for the set and classify them with LR.  When the sample without a given token approaches the
hyperplane, it means that the token was relevant to the class defined by the plane.
We gathered all of the distances for every document in a set and then
normalized them using min-max scaling. The final score is a difference of the obtained
distance from the mean value. The color intensity in Figures
\ref{fig:idsinterp-csic2010} and \ref{fig:idsinterp-ids2018} represents those
scores for every token (we coloured only tokens positively correlated with
anomalies).

In Table \ref{tab:importance} we present top 24 features for two datasets (sum across 50
different documents). For CSIC2010 we decided to generate the table for
neighbourhood of the sample mentioned in Table \ref{tab:simmilar}. Interestingly, the
``\textbackslash'' token has the highest score in general. This is because it is
related with CRLF injection attacks (negatively correlated tokens must
balance this feature). The rest of the table shows token highly related with
SQL Injection.

In the case of CSE-CIS-IDS2018 dataset, we took 50 random samples to determine
why the dataset is so easy to classify. As you can see (see also Figure
\ref{fig:idsinterp-ids2018}), the ``Accept'' and ``Accept-Encoding'' are highly related to anomaly
subset. This means that those two lines appear in almost every sample in nearly
unchanged form (this is a very strong feature). This is why we obtained such
excellent results. We had a similar issue with CSIC2010.
It turned out that each anomalous request ended with a ``\textbackslash n'' character and not ``\textbackslash r \textbackslash n''. It is intriguing how many other researchers also faced this issue.

\begin{table}[htb!]
\small
\centering
\caption{The top 24 features in anomalous traffic. For IDS2018 we took 50 randomly sampled request. For CSIC2021, we took 50 requests closest to a given sample from Table \ref{tab:simmilar}}
\label{tab:importance}
\begin{tabular}{cccc}
\hline
\multicolumn{4}{c}{CSE-CIS-IDS2018} \\ \hline\hline
token        & score & token  & score \\ \hline
-           & 18.56 & html  & 5.29    \\ \hline
/           & 16.33 & xhtml & 5.24    \\ \hline
Accept      & 14.32 & ;     & 5.09    \\ \hline
:           & 11.41 & +     & 4.81    \\ \hline
xml         & 10.74 & 0     & 4.15    \\ \hline
,           & 9.55  & text  & 3.19    \\ \hline
Encoding    & 9.13  & 9     & 2.59    \\ \hline
deflate     & 7.78  & =     & 2.30    \\ \hline
.           & 7.50  & 8     & 2.13    \\ \hline
gzip        & 7.42  & GET   & 2.11    \\ \hline
application & 7.19  & login & 1.94    \\ \hline
q           & 6.85  & php   & 1.43    \\ \hline
\end{tabular}
\hspace{2ex}
\begin{tabular}{cccc}
\hline
\multicolumn{4}{c}{CSIC2010}                \\ \hline\hline
token            & score  & token   & score \\ \hline
\textbackslash{} & 203.38 & 5      & 17.20 \\ \hline
r                & 174.68 & 3      & 17.10 \\ \hline
+                & 68.24  & LIKE   & 12.28 \\ \hline
\%               & 41.09  & +\%    & 11.69 \\ \hline
.                & 36.05  & datos  & 11.58 \\ \hline
=                & 32.34  & +*+    & 11.28 \\ \hline
/                & 31.32  & FROM   & 11.14 \\ \hline
n                & 28.86  & TABLE  & 11.09 \\ \hline
1                & 23.78  & WHERE  & 11.04 \\ \hline
B                & 22.58  & DROP   & 10.57 \\ \hline
27               & 17.75  & \&     & 10.54 \\ \hline
0                & 17.30  & SELECT & 10.21 \\ \hline
\end{tabular}

\end{table}

\begin{table}[htb!]
\tiny
\centering
\caption{Few n-th closest samples to the given one (sample number ''0'').}
\label{tab:simmilar}
\begin{tabular}{cp{0.85\textwidth}}
\hline
N-th closest\\ neighbour & \multicolumn{1}{c}{Datasets} \\ \hline
0   & \url{/vaciar.jsp?B2=Vaciar+carrito\%27\%3B+DROP+TABLE+usuarios\%3B+SELECT+*+FROM+datos+WHERE+nombre+LIKE+\%27\%25}                    \\ \hline
1   & \url{/vaciar.jsp?B2=Vaciar+carrito\%27\%3B+DROP+TABLE+usuarios\%3B+SELECT+*+FROM+datos+WHERE+nombre+LIKE+\%27\%25}                    \\ \hline
5   & \url{/vaciar.jsp?B2=Vaciar+carrito\%27\%3B+DROP+TABLE+usuarios\%3B+SELECT+*+FROM+datos+WHERE+nombre+LIKE+\%27\%25}                    \\ \hline
10  & \url{/vaciar.jsp?B2=\%27\%3B+DROP+TABLE+usuarios\%3B+SELECT+*+FROM+datos+WHERE+nombre+LIKE+\%27\%25}                                  \\ \hline
15  & \url{/entrar.jsp?errorMsg=\%27\%3B+DROP+TABLE+usuarios\%3B+SELECT+*+FROM+datos+WHERE+nombre+LIKE+\%27\%25}                            \\ \hline
25  & \url{/entrar.jsp?errorMsg=Credenciales+incorrectas\%27\%3B+DROP+TABLE+usuarios\%3B+SELECT+*+FROM+datos+WHERE+nombre+LIKE+\%27\%25}    \\ \hline
50  & \url{/anadir.jsp?id=2\&nombre=Jam\%F3n+Ib\%E9rico\&precio=85\&cantidad=\%27\%3B+DROP+TABLE+usuarios\%3B+SELECT+*+FROM+datos+WHERE+nombre+LIKE+\%27\%25\&B1=A\%F1adir+al+carrito}  \\ \hline
100 & \url{/anadir.jsp?id=3\&nombre=Queso+Manchego\&precio=39\&cantidad=86\&B1=\%3C\%21--\%23exec+cmd\%3D\%22rm+-rf+\%2F\%3Bcat+\%2Fetc\%2Fpasswd\%22+--\%3E}                           \\ \hline
200 & \url{/anadir.jsp?id=2\&nombre=Queso+Manchego\&precio=85\&cantidad=86\%22\%3E\%3C\%21--\%23EXEC+cmd\%3D\%22dir+\%22--\%3E\%3C\&B1=A\%F1adir+al+carrito}                            \\ \hline
\end{tabular}
\end{table}

\section{Conclusions}
\label{section:conslusions}

We proposed a method for analyzing anomalous HTTP traffic that uses vector representations of HTTP requests based on the RoBERTa model -- a state-of-the-art text representation technique. Using these representations, we can successfully discriminate normal and anomalous HTTP packets: we have shown the effectiveness of the representations in the supervised classification task using CSIC2010 and CSE-CIC-IDS2018 datasets and obtained SOTA accuracy on these benchmarks. We showed that these models generalize to new data, as we successfully discovered anomalies in real HTTP traffic in a dataset we curated (and make available as a supplementary asset to this work).     

The important characteristic of the proposed method is that the vector representations allow us to analyze anomalous HTTP requests in terms of interpretable subsets of tokens/features -- we showed how to obtain such informative patterns. 

In addition to this, we observe that vectorized HTTP requests tend to group into clear, disjoint clusters of similar, normal, or anomalous requests. This suggests that in addition to supervised anomaly detection, unsupervised methods of anomalous traffic are likely to be effective -- we want to address this in our future work.  
We based our research only on data collected in pcap format through a network traffic logger. We have not used HTTP server access and error logs. We believe that these can be easily correlated with the analyzed data and even be used for automated attack classification -- which forms another direction for further work.

\textbf{Limitations} and directions for future research. 

The size of the language model is large (dimensionality of the vector representation is ca.\ 3000), which affects the training and inference time (the training process for CSIC2010 dataset took about 7 hours using two GPUs RTX 2080Ti). Further research is needed to reduce the size of the language model \slash{} reduce the dimensionality of vector representation to simplify downstream supervised\slash{}unsupervised analysis. 

In this work, we used supervised methods of HTTP anomaly detection. This approach requires that annotated dataset is available for model training (which may be expensive to curate) and allows us to discover only attack types known in the training data. These limitations can be mitigated by using unsupervised anomaly detection techniques -- the observations we make in this work suggest that this may be effective using the vector representations.   
   
Finally, a clear limitation of this and similar works comes from the limited availability of modern, representative network traffic datasets. Most of the research is based only on one CSIC2010 dataset and other self-prepared data (collected by hand and not published or extracted from pcaps). The quality of such datasets is hard to determine and may be questionable, which affects the quality/generalization of traffic models.

\bibliographystyle{plainnat}
\bibliography{main.bib}

\end{document}